\documentclass{article}

\PassOptionsToPackage{numbers,square,comma}{natbib}

\usepackage[preprint]{nips_2018}

\usepackage[utf8]{inputenc} 
\usepackage[T1]{fontenc}    
\usepackage{hyperref}       
\usepackage{url}            
\usepackage{booktabs}       
\usepackage{amsfonts}       
\usepackage{nicefrac}       
\usepackage{microtype}      
\usepackage{graphicx}
\usepackage{breqn}
\usepackage{multirow}
\usepackage[toc,page]{appendix}
\usepackage{rotating}

\newcommand{\argmin}{\mathop{\mathrm{arg\,min}}}

\title{Controllable Generative Adversarial Network}

\author{
  Minhyeok Lee \\
  School of Electrical Engineering \\
  Korea University \\
  Seoul, Korea 02841 \\
  \texttt{suam6409@korea.ac.kr} \\
  \And
  Junhee Seok \thanks{To whom correspondence should be addressed} \\
  School of Electrical Engineering \\
  Korea University \\
  Seoul, Korea 02841 \\
  \texttt{jseok14@korea.ac.kr} \\
}

\begin{document}

\maketitle

\begin{abstract}
Recently introduced generative adversarial network (GAN) has been shown numerous promising results to generate realistic samples. The essential task of GAN is to control the features of samples generated from a random distribution. While the current GAN structures, such as conditional GAN, successfully generate samples with desired major features, they often fail to produce detailed features that bring specific differences among samples. To overcome this limitation, here we propose a controllable GAN (ControlGAN) structure. By separating a feature classifier from a discriminator, the generator of ControlGAN is designed to learn generating synthetic samples with the specific detailed features. Evaluated with multiple image datasets, ControlGAN shows a power to generate improved samples with well-controlled features. Furthermore, we demonstrate that ControlGAN can generate intermediate features and opposite features for interpolated and extrapolated input labels that are not used in the training process. It implies that ControlGAN can significantly contribute to the variety of generated samples.
\end{abstract}

\section{Introduction}
Generative Adversarial Network (GAN) is a neural network structure, which has been introduced for generating realistic samples. GAN consists of two modules, a generator and a discriminator. A generator produces fake samples from random noises, while a discriminator attempts to distinguish between these fake samples and real samples. The generator tries to deceive the discriminator by learning from errors which are the output of discriminator with fake samples. By such an adversarial and competitive learning, latent variables of the samples are mapped onto random variables which are the input of generators. After adequate learning iterations of such a process, the generator can generate realistic samples from random noises.

While it is introduced recently, GAN has shown many promising results not only for generating realistic samples \citep{r1,r2,r3}, but also for machine translation \citep{r4} and image super-resolution \citep{r5}.

However, for generating realistic samples, we can hardly control the GAN because the random distribution is used for the input variables of generators. While vanilla GAN can generate realistic samples from a random noise, the relationship between inputs of the generator and features of generated samples is not obvious. In the last few years, there have been several attempts to control generated samples by GAN \citep{r6,r7,r8,r9,r10}. One of the most popular methods to control GAN is the conditional GAN \citep{r10}. Conditional GAN inputs labels into the generator and the discriminator so that they work under the conditions.

The current conditional GAN mainly focuses on generating realistic samples, rather than making difference between generated samples according to input labels. As results, it is difficult to generate samples with detailed features while conditional GAN is successful for major features. For example, with CelebA dataset \citep{r11} which consists of 202,559 celebrity face images labeled with 40 different features, conditional GAN only works with major features, such as smiling or bangs. In order to make the generator work with detailed features, such as pointy nose or arched eyebrows, we have to control the generator to more focus on generating different samples according to input labels.

In this paper, we propose a novel architecture of the generative model to control generated samples, called Controllable Generative Adversarial Network (ControlGAN). ControlGAN is composed of three players, a generator/decoder, a discriminator and a classifier/encoder. The generator in ControlGAN plays the games with the discriminator and the classifier simultaneously in our method; the generator aims to deceive the discriminator and be classified correctly by the classifier.

ControlGAN has two main advantages compared to existing models. First, ControlGAN can be trained to focus more on input labels so that ControlGAN can generate samples with detailed feature where conditional GAN can hardly generate. Second, ControlGAN uses an independent network for mapping the features into corresponding input labels while the discriminator conducts such a work in conditional GAN and other conditional variants of GAN \citep{acgan}. Consequently, the discriminator can more concentrate on its own objective, which is the discrimination between fake samples and original samples, so that the quality of generated samples can be enhanced.

\begin{figure}
\centering
\includegraphics[width=0.98\textwidth]{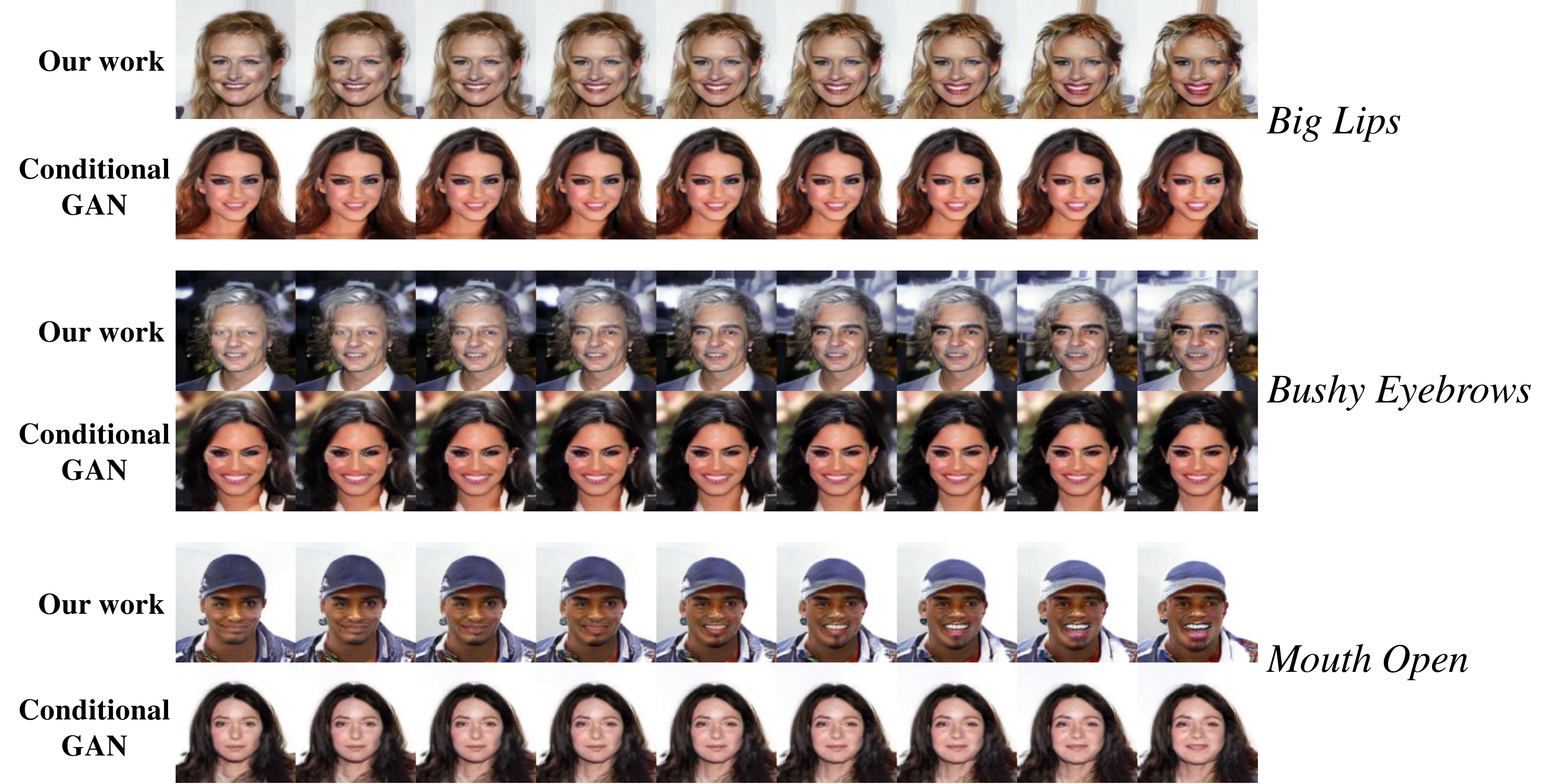}
\caption{\textbf{Comparison between ControlGAN and conditional GAN for generating face images with detailed features.} Each row is generated with the same input noise $z$. The images in the right most column are generated with $3 \times label$, and the images in the left most columns are generated with $-1 \times label$. The intermediate images are generated with interpolated and extrapolated label values}
\label{fig:fig_intro}
\end{figure}

ControlGAN is applied to the CelebA dataset and LSUN dataset \citep{lsun} in this paper. As shown in Figure \ref{fig:fig_intro}, we demonstrate that ControlGAN can effectively generate face images according to input labels. Furthermore, we demonstrate the ControlGAN also works with extrapolated label values. To evaluate such a zero-shot learning of ControlGAN, we test the ControlGAN with untrained label values in Section \ref{section_ev}.

\section{Background}
\subsection{A brief review of generative adversarial networks and its conditional variants}
GAN is a neural network structure for learning to generate samples and mapping latent variables of a dataset. Given a dataset $X=\{x_1,...,x_n\}$ where $x \in \Re^k$, if the samples are not orthogonal to each other, latent variables $z \in \Re^{l<k}$ exist. For example, for a face image dataset, latent variables can be the attribute of human faces, such as shape of face, sharpness of nose, and color of eyes.

Auto-encoder(AE) is a neural network structure to encode samples into latent variables to decrease dimensionality of a dataset, i.e. $f_{AE} \in \Re^{k} \rightarrow \Re^{l}$. The objective of generative models is the inverse function of the AE ($f_{GM} \in \Re^{l} \rightarrow \Re^{k}$), which means, given latent variables, the models aim to generate samples. Therefore, the structure of generative models using neural network architecture is similar to the decoder of AE. The problem of generative models is how to find these latent variables and learn to generate samples.

GAN solves such a problem by a competitive learning process between the generator and the discriminator. First, a generator generates samples from randomly initialized variables. Then, a discriminator learns to distinguish between the generated samples and real samples. Simultaneously, the generator learns to deceive the discriminator by losses of the generated samples. By repeating such a process, the generator can learn to generate realistic samples and embed the latent variables into input variables of the generator.

However, since the relation between generated samples and input variables is not obvious, the generated samples cannot be controlled as we desire to. For example, we cannot control GAN to generate face image samples of a smiling old woman having blond hair; we have to select the face images from randomly generated samples when we use vanilla GAN. To address such a problem, conditional variants of GAN have been studied.

Conditional GAN is the most popular GAN structure to control the generated samples from a generator. Conditional GAN takes label inputs for the generator and the discriminator to work the generator under the condition of the input labels.

Several studies have been conducted for using a classifier to address the problem. Auxiliary Classifier GAN (AC-GAN) \citep{acgan} uses a classifier as the discriminator of GAN structure. Triple-GAN \citep{triplegan} uses the classification results as an input for discriminator. However, such methods commonly use a classifier that is attached to a discriminator. Therefore, since the discriminator decides the condition of samples, the methods can hardly handle the limitation of conditional GAN. 

\subsection{The limitation of conditional GAN}
\begin{figure}
\centering
\includegraphics[width=0.98\textwidth]{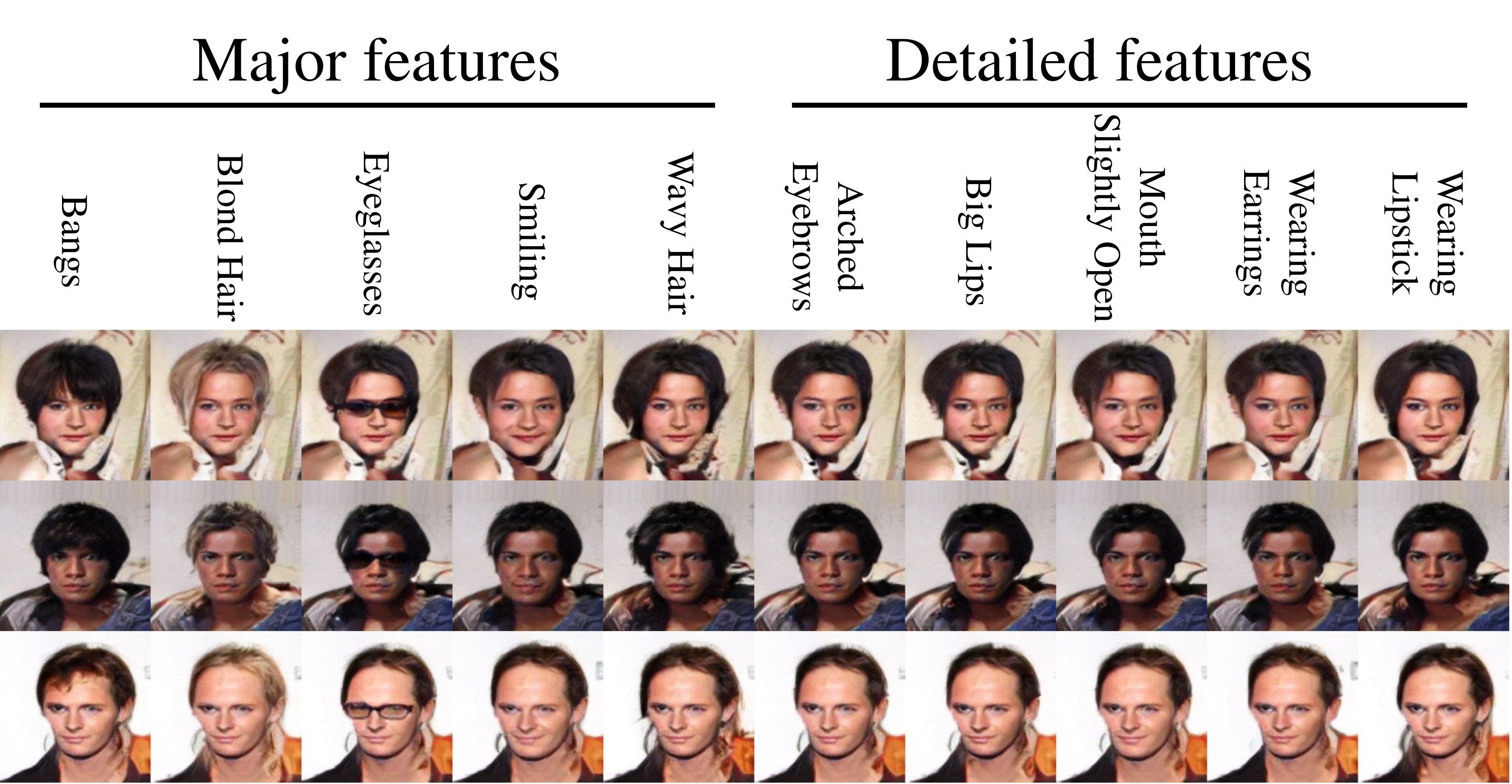}
\caption{\textbf{Generated face images by conditional GAN with ten different labels.}}
\label{fig:fig_lim}
\end{figure}

While the conditional GAN is the most popular GAN structure to generate conditional samples, conditional GAN frequently fails to generate detailed features. For example, while conditional GAN can generate a face image sample with a condition that is easily distinguishable, such as 'Blond Hair', conditional GAN can hardly generate face images with some detailed labels such as 'Arched Eyebrows', 'Big Lips', 'Mouth Slightly Open', 'Wearing Earrings' and 'Wearing Lipstick', as shown in Figure \ref{fig:fig_lim}.

Such a failure occurs because the discriminator decides whether the label/condition is correct. The main objective of discriminator is the discrimination between fake and real samples. Therefore, if a condition (or a label) is very rare in a dataset or is far from the center of sample distribution where the samples densely exist, the probability that the discriminator decides the samples with such conditions are fake samples increases.

\section{Methods}
\subsection{Controllable generative adversarial networks}
\begin{figure}
\centering
\includegraphics[width=0.98\textwidth]{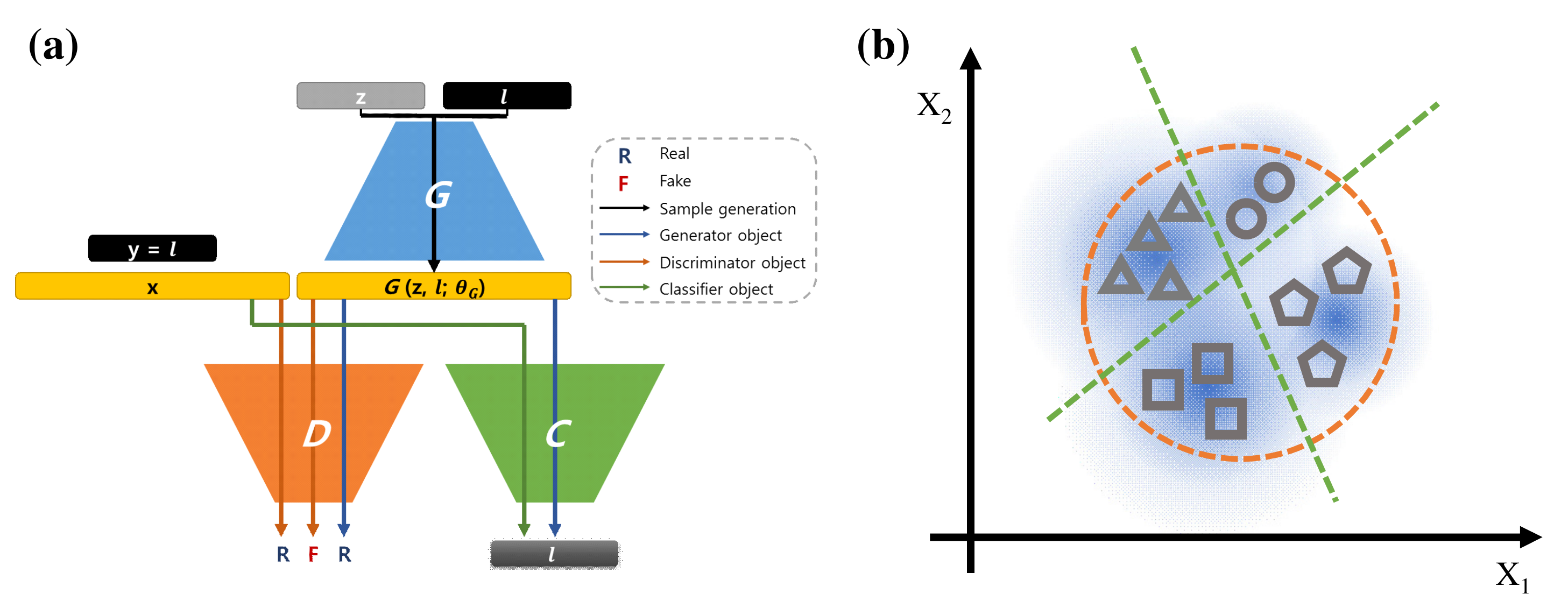}
\caption{\textbf{The concept of ControlGAN. (a)} The architecture of ControlGAN. \textbf{(b)} An illustration of the concept of ControlGAN. The green dashed line denotes the classifier and the orange dashed line denotes the discriminator. The grey figures denote samples labeled with different class. The generator (blue region) tries to learn the sample distribution and be classified to correct labels, simultaneously.}
\label{fig:fig1}
\end{figure}

ControlGAN is composed of three neural network structures, which are a generator/decoder, a discriminator and a classifier/encoder. Figure \ref{fig:fig1} illustrates the architecture of ControlGAN. Three-player game is conducted in ControlGAN where the generator tries to deceive the discriminator, which is the same as vanilla GAN, and simultaneously aim to be classified corresponding class by the classifier. The generator and the classifier can be interpreted as a decoder-encoder structure because labels are commonly used for inputs for the generator and outputs for the classifier.

ControlGAN minimizes the following equations:

\begin{dmath}
\theta_{D}=\argmin \{ \alpha \cdot L_{D} (t_{D},D(x;\theta_{D})) + (1-\alpha) \cdot L_{D}((1-t_{D}),D(G(z,l;\theta_{G});\theta_{D})) \}, 
\end{dmath}
\begin{dmath}
\theta_{G}=\argmin \{ \gamma_{t} \cdot L_{C} (l,G(z,l;\theta_{G})) + L_{D} (t_{D},D(G(z,l;\theta_{G});\theta_{D})) \},
\end{dmath}
\begin{dmath}
\theta_{C}=\argmin \{ L_{C} (l,x;\theta_{C}) \},
\end{dmath}

where $l$ is the binary representation of labels of sample $x$ and input data for the generator, $t_{D}$ is the label for discriminator which we set to one in this work, and $\alpha$ denotes a parameter for the discriminator.

ControlGAN forces features to be mapped onto corresponding $l$ inputted into the generator. The parameter $\gamma_{t}$ decides how much the generator focus on the input labels for the generator.

It is important to maintain the equilibrium between the two objectives of the generator in ControlGAN since the ControlGAN aims to optimize a decoder-encoder structure and a GAN structure simultaneously. Suppose a well-trained conditional generator $G(z, l)$ which perfectly learned a true distribution exists, then a set of generated samples from the generator has same classification loss with the original dataset:

\begin{dmath}
E=\frac {L_{C} (l,G(z,l;\theta_{G}))}{L_{C} (l,x)}
 = 1 \hspace{0.3cm} if \hspace{0.2cm} G(z,l) \hiderel{=} P(X)
\label{eq:EQC}
\end{dmath}

If a generator is trained to concentrate on the input labels, the value $E$ in (\ref{eq:EQC}) would be less than one, and otherwise the value would be more than one.

ControlGAN controls whether to concentrate on learning the distribution of a dataset or learning to generate samples according to input labels by the parameter $E$ that maintains the classification loss of generated samples constantly. The $\gamma_{t}$ is a learning parameter to maintain $E$ which is changed by time step $t$, i.e. the iteration process, and is calculated as follows:

\begin{dmath}
\gamma_{t}=\gamma_{t-1} + r\cdot\{L_{C} (l,G(z,l;\theta_{G})) - E\cdot L_{C} (l,x) \},
\label{eq:gamma}
\end{dmath}

where $r$ is a learning rate parameter for $\gamma_{t}$.

Such a concept of the equilibrium parameter is similar to that of Boundary Equilibrium GAN (BEGAN) \citep{r3}. BEGAN employs an equilibrium parameter to maintain the balance between the generator and the discriminator. In this work, the equilibrium parameter is used for the balance between the learning of the GAN structure and the decoder-encoder structure.

\subsection{ControlDCGAN structure for the applications}

In this work, we used ControlDCGAN structure which is a combination of the architecture of ControlGAN and Deep Convolutional Generative Adversarial Network (DCGAN) \citep{dcgan}. We used the residual modules for the generator, the discriminator and the classifier. The batch normalization or dropout is not used to evaluate the vanilla ControlGAN. The generator, the discriminator and the classifier consist of 19, 22 and 22 hidden layers, respectively. The size of generated samples from the generator is $128 \times 128 \times 3$. The architecture we used for the application is summarized in Table \ref{arch}.

\begin{table}
  \caption{\textbf{The architecture of each module of ControlDCGAN for the applications.} $s$ denotes the stride for convolutional or deconvolutional layers.}
  \label{arch}
  \centering
  \qquad
  \setlength\tabcolsep{1.5pt}
  \begin{tabular}{l | c c c c c c}
    \toprule
    \textbf{Description} & \multicolumn{2}{c}{\textbf{Generator}}  & \multicolumn{2}{c}{\textbf{Discriminator}} & \multicolumn{2}{c}{\textbf{Classifier}} \\
    \midrule
    Input       & Concatenate($z$, $l$) &  & $x$ or $G(z,l)$  &  & $x$ or $G(z,l)$ & \\
    \midrule
    Fully connected & FC($32\times 32 \times 64$) &    & None  & & None  & \\
    Convolutional $(s=2)$ & None &    & Conv(5, 5, 64)  & & Conv(5, 5, 64) & \\
    \midrule
    \multirow{ 2}{*}{Residual module 1} & Deconv(3, 3, 64) & \multirow{ 2}{*}{$ \times 2$}& Conv(3, 3, 64) & \multirow{ 2}{*}{$ \times 2$} & Conv(3, 3, 64)& \multirow{ 2}{*}{$ \times 2$}\\
                                        & Deconv(3, 3, 64) & & Conv(3, 3, 64) & & Conv(3, 3, 64)& \\
    \midrule
    Deconvolutional $(s=2)$ & Deconv(5, 5, 64) & & None & & None & \\
    Pooling & None & & AveragePool(2, 2) & & AveragePool(2, 2) & \\
    \midrule
    \multirow{ 2}{*}{Residual module 2} & Deconv(3, 3, 64) & \multirow{ 2}{*}{$ \times 4$}& Conv(3, 3, 64) & \multirow{ 2}{*}{$ \times 4$} & Conv(3, 3, 64)& \multirow{ 2}{*}{$ \times 4$}\\
                                        & Deconv(3, 3, 64) & & Conv(3, 3, 64) & & Conv(3, 3, 64)& \\
    \midrule
    Deconvolutional $(s=2)$ & Deconv(5, 5, 64) & & None & & None & \\
    Pooling & None & & AveragePool(2, 2) & & AveragePool(2, 2) & \\
    \midrule
    \multirow{ 2}{*}{Residual module 3} & Deconv(3, 3, 64) & \multirow{ 2}{*}{$ \times 2$}& Conv(3, 3, 64) & \multirow{ 2}{*}{$ \times 4$} & Conv(3, 3, 64)& \multirow{ 2}{*}{$ \times 4$}\\
                                        & Deconv(3, 3, 64) & & Conv(3, 3, 64) & & Conv(3, 3, 64)& \\
    \midrule
    Deconvolutional $(s=1)$& Deconv(5, 5, 64) & & None & & None & \\
    Pooling & None & & AveragePool(2, 2) & & AveragePool(2, 2) & \\
    Fully connected & None & & FC(128) & & FC(128) & \\
    Fully connected & None & & FC(1) & & FC(Num. of classes) & \\

    \bottomrule
  \end{tabular}
\end{table}

\section{Results and Discussion}
\subsection{Generating multi-label image samples of celebrity face using CelebA dataset}
In this section, ControlGAN was trained over the CelebA dataset \citep{r11}. The CelebA dataset contains celebrity face images with multiple labels for each image. For example, a sample can have multiple labels of `Attractive', `Blond Hair', `Mouth Slightly Open' and `Smiling'.

We used Adam optimizer with learning rate of $2 \times 10^{-4}$ and $5 \times 10^{-5}$ to train the model. The learning rate decreases after 30 epochs, and the model was trained 20 more epochs with the decreased learning rate. The equilibrium parameter $E$ was set to 0.05, 0.5 and 1.0. The learning rate parameter $r$ was set to 0.01, and $\alpha$ was set to 0.5. As for the inputs for the generator, the 500-dimensional uniform distribution, i.e. $Unif(-1,1)$, and the binary encoded labels were employed. We used leaky ReLU activation $(\alpha = 0.1)$ function for the generator, the discriminator and the classifier.

A pre-training process was conducted for the classifier instead of a simultaneous training of the GAN structure and the decoder-encoder structure. After the two epochs of pre-training, the classifier was fixed and no more training had been conducted during the GAN structure training process.

\begin{figure}
\centering
\includegraphics[width=0.98\textwidth]{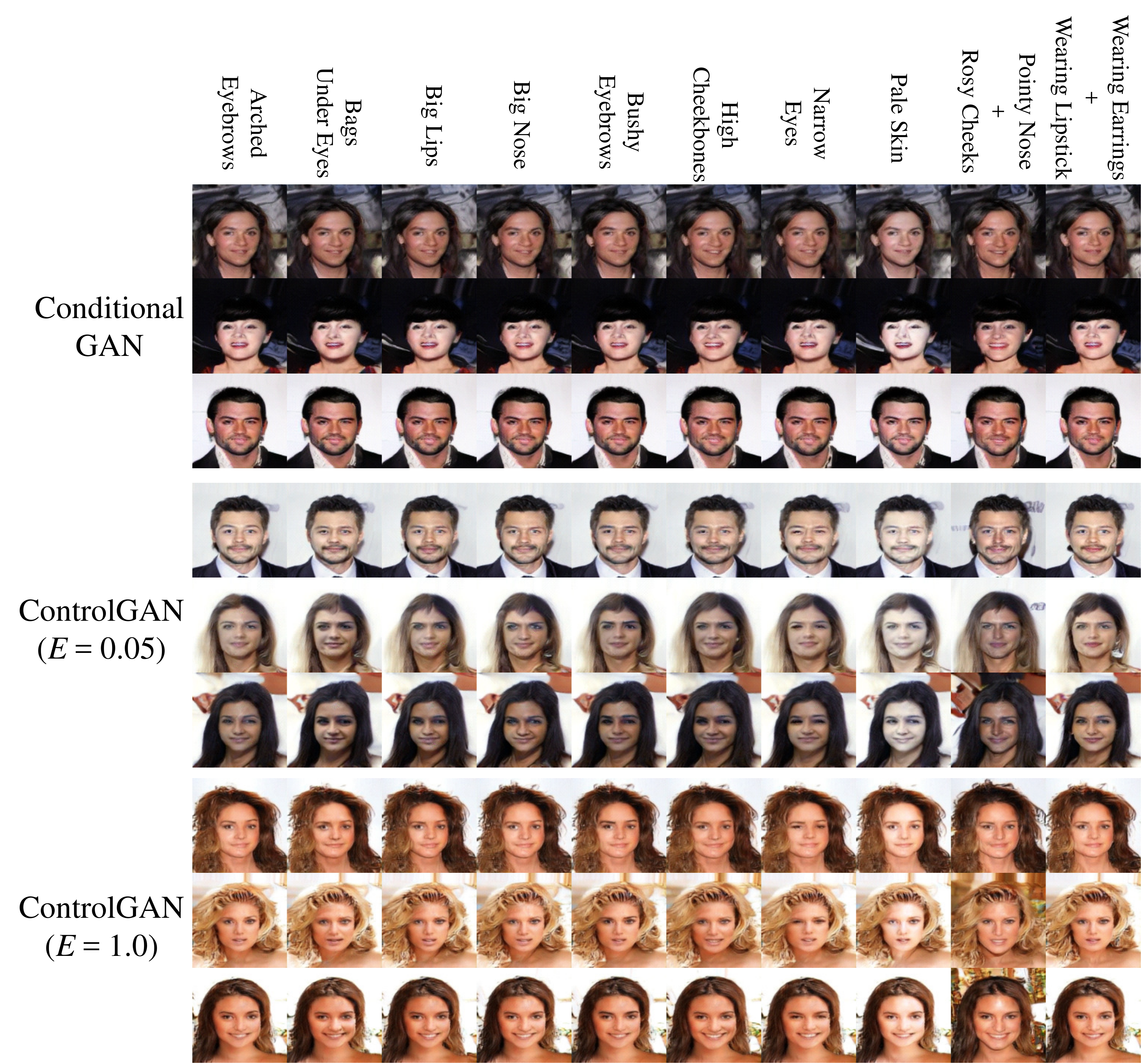}
\caption{\textbf{Comparison between conditional GAN and ControlGAN.} Images in each row are generated with a same input noise $z$. Each column denotes the corresponding labels. Two labels are used together in the last two column. All images are the size of $128 \times 128$.}
\label{fig:fig_comp}
\end{figure}

Figure \ref{fig:fig_comp} is a comparison between the ControlGAN and the conditional GAN. As shown in Figure \ref{fig:fig_comp}, there are very little differences between labels/columns in condtional GAN. Generally, generated face images by ControlGAN follows the input labels well compared to the conditional GAN. For example, with the `Arched Eyebrows' label, all image samples generated by ControlGAN follows the label while the face images generated by conditional GAN hardly show the difference.

As we described in the previous section, ControlGAN has an advantage for generating label-focused samples by choosing a low value of $E$. By selecting a low value of $E$, which means the generated samples have a low value of classification error, we can make the generator more focus on the input labels. Figure \ref{fig:fig_comp} shows the comparison between ControlGAN with different $E$. As shown in Figure \ref{fig:fig_comp}, with low $E$, the generated images are significantly label-focused. In $E=0.05$, the images generated with the condition of 'Pale Skin', correspond to an unreal level where similar genuine samples rarely or do not exist in the CelebA dataset.

Such a property is the main advantage of ControlGAN since it proves that ControlGAN can generate samples beyond the training set. We will describe further such a zero-shot property of ControlGAN in Section \ref{section_ev}

\subsection{Room image generation with LSUN dataset}
In this section, in order to demonstrate generalizability of ControlGAN, ControlGAN was trained with a different dataset, which is a large scale scene dataset called LSUN \citep{lsun}. The dataset consists of ten different places, such as a bedroom and a restaurant. Among the places, we selected four different labels corresponding to indoor house rooms, i.e. 'Bedroom', 'Dining room', 'Kitchen' and 'Living room'.

\begin{figure}
\centering
\includegraphics[width=0.98\textwidth]{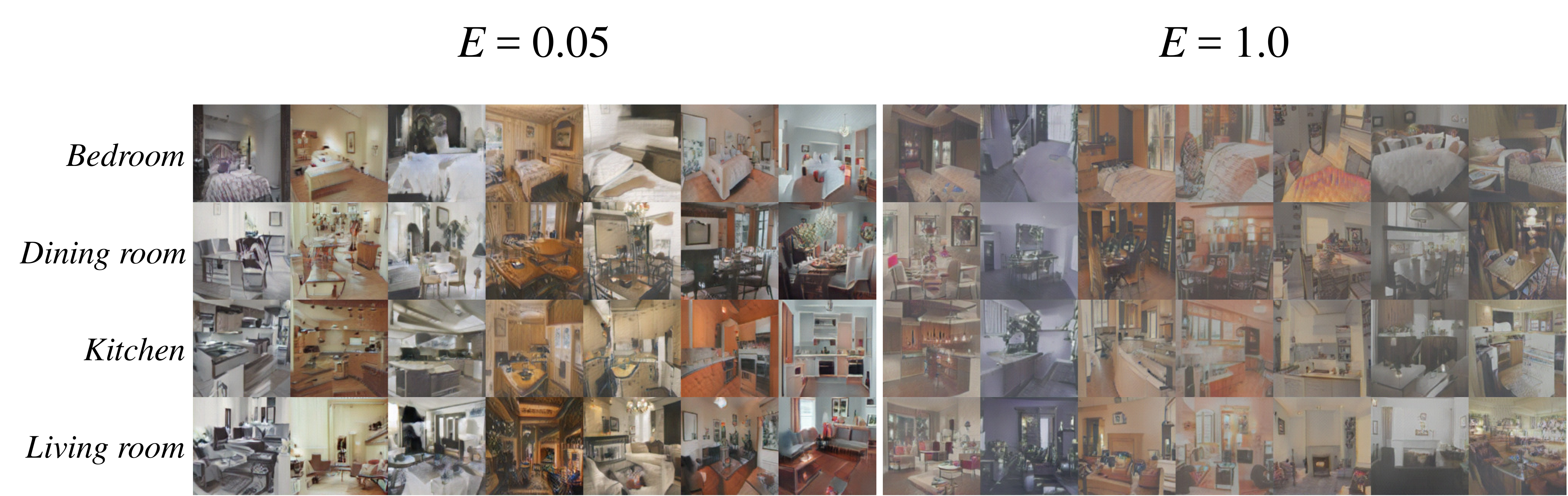}
\caption{\textbf{$\mathbf{128 \times 128}$ size of conditional room images generated by ControlGAN.}}
\label{fig:fig_lsun}
\end{figure}

The architecture we used for the application is the same as the structure of the previous section. We used the learning rate of $5 \times 10^{-5}$, and $E$ is set to 0.05, 0.5 and 1.0. A pre-trained classifier was also used. The classifier was trained for 0.1 epoch. The generator and discriminator were trained for one epoch.

As shown in Figure \ref{fig:fig_lsun}, ControlGAN can learn the features of each room, and successfully generate room images according to the labels. According to our expectation, the equilibrium parameter $E$ decides the degree to concentrate on the input labels as same as the previous application. However, we found that ControlGAN can generate much clearer images with a low $E=0.05$. Such a property is conjectured because the classifier can assist the GAN training by forcibly mapping the features of labels onto the input $l$, as we described in Section 2 and Figure \ref{fig:fig1}.

\subsection{Interpolation \& extrapolation of labels}
\label{section_ev}
In order to demonstrate that the ControlGAN learns the features and does not just memorize the training set, the input labels are interpolated and extrapolated in this section. Note that the labels were one-hot or binary encoded in the training process, therefore the interpolated values had never been trained.

For a further demonstration of the effectiveness of ControlGAN, we used the extrapolated values between [-1.0, 3.0]. Since the input labels were binary or one-hot encoded, and the training had been conducted with only the two values of 0.0 and 1.0, we expected to obtain a half-smile face image with the value of 0.5 for the 'Smiling' label and a perfect-smile face image with the value of 3.0 if the ControlGAN learns the features well.

\begin{figure}
\centering
\includegraphics[width=0.98\textwidth]{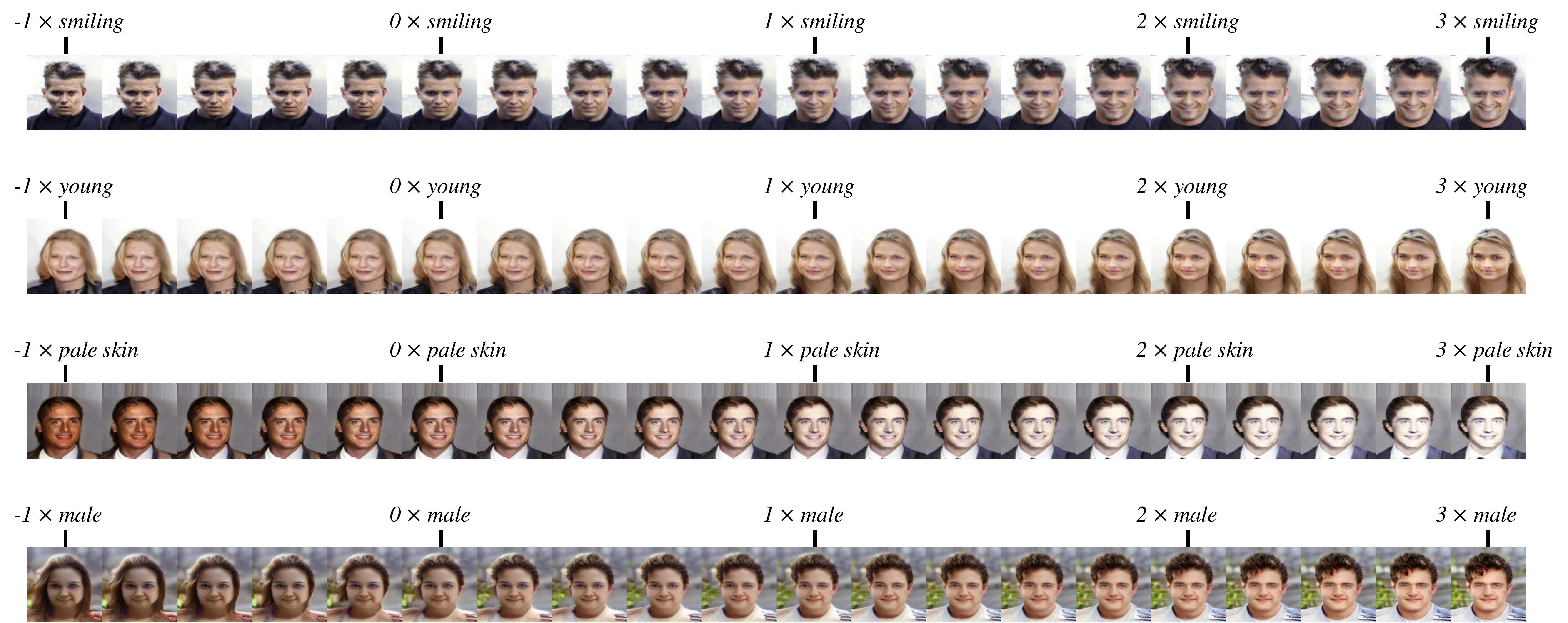}
\caption{\textbf{Interpolation and extrapolation of labels.} Note that the model was trained only with $0 \times label$ and $1 \times label$. The values of $[-1.0, 0.0)$, $(0.0, 1.0)$ and $(1.0, 3.0]$ in this figure have never been trained in the training process. $E$ is set to 0.05.}
\label{fig:fig_inter}
\end{figure}

As shown in Figure \ref{fig:fig_inter}, ControlGAN conducts well with interpolated and extrapolated values of labels. Interestingly, the generated face image with $-1 \times smiling$ label corresponds to a frown or angry face, which is an untrained feature in the training process. Likewise, $-1 \times pale skin$ corresponds to the dark skin which is not contained in the labels of CelebA dataset.

Such a result implies the ControlGAN can learn the attributes of input labels. One can easily conjecture that the opposite of a smiling face might be a frown or angry face; ControlGAN can do such a conjecture as well while it have never been trained.  

\section{Conclusion}
In this paper, we proposed a generative model, called ControlGAN, which can effectively control generated samples. ControlGAN consists of three modules, i.e. a generator/decoder, a discriminator and a classifier/encoder. By mapping corresponding features into input labels, generated samples can be controlled according to the labels.

While ControlGAN is a simple architecture, which is a combination of the vanilla GAN and a decoder-encoder structure, we demonstrated that ControlGAN works well to generate conditional samples. We employed DCGAN architecture to demonstrate the applications; however, the quality of samples can be enhanced with a state-of-art GAN structure, such as StackGAN \citep{r1}, WGAN \citep{wgan} and BEGAN \citep{r3}.

Furthermore, we demonstrated the ControlGAN conducts with zero-shot values, by feeding interpolated and extrapolated values to the generator. Since the proposed architecture shows powerful performance to control generated samples, we expect that ControlGAN can contribute to the research in generative models.

\medskip

\small

\bibliographystyle{ieeetr}
\bibliography{CGAN}

\newpage

\appendix
\normalsize
\section{List of Appendix}
Figure \ref{fig:fig_comp_add}: \textbf{Comparison between conditional GAN and ControlGAN with $E=0.05$, $0.5$ and $1.0$.}

Figure \ref{fig:fig_lsun_add}: \textbf{$128 \times 128$ size of conditional room images generated by ControlGAN.}

Figure \ref{fig:fig_inter_add1}: \textbf{Interpolation of the 'Smiling', 'Young', 'Pale Skin' and 'Male' labels.}

Figure \ref{fig:fig_inter_add2}: \textbf{Interpolation of the 'Attractive', 'Big Lips', 'Big Nose' and 'Bushy Eyebrows' labels.}

Figure \ref{fig:fig_inter_add3}: \textbf{Interpolation of the 'Chubby', 'Heavy Makeup', 'Mouth Slightly Open' and 'Pointy Nose' labels.}

\begin{sidewaysfigure}
    \centering
    \includegraphics[width=0.95\textwidth]{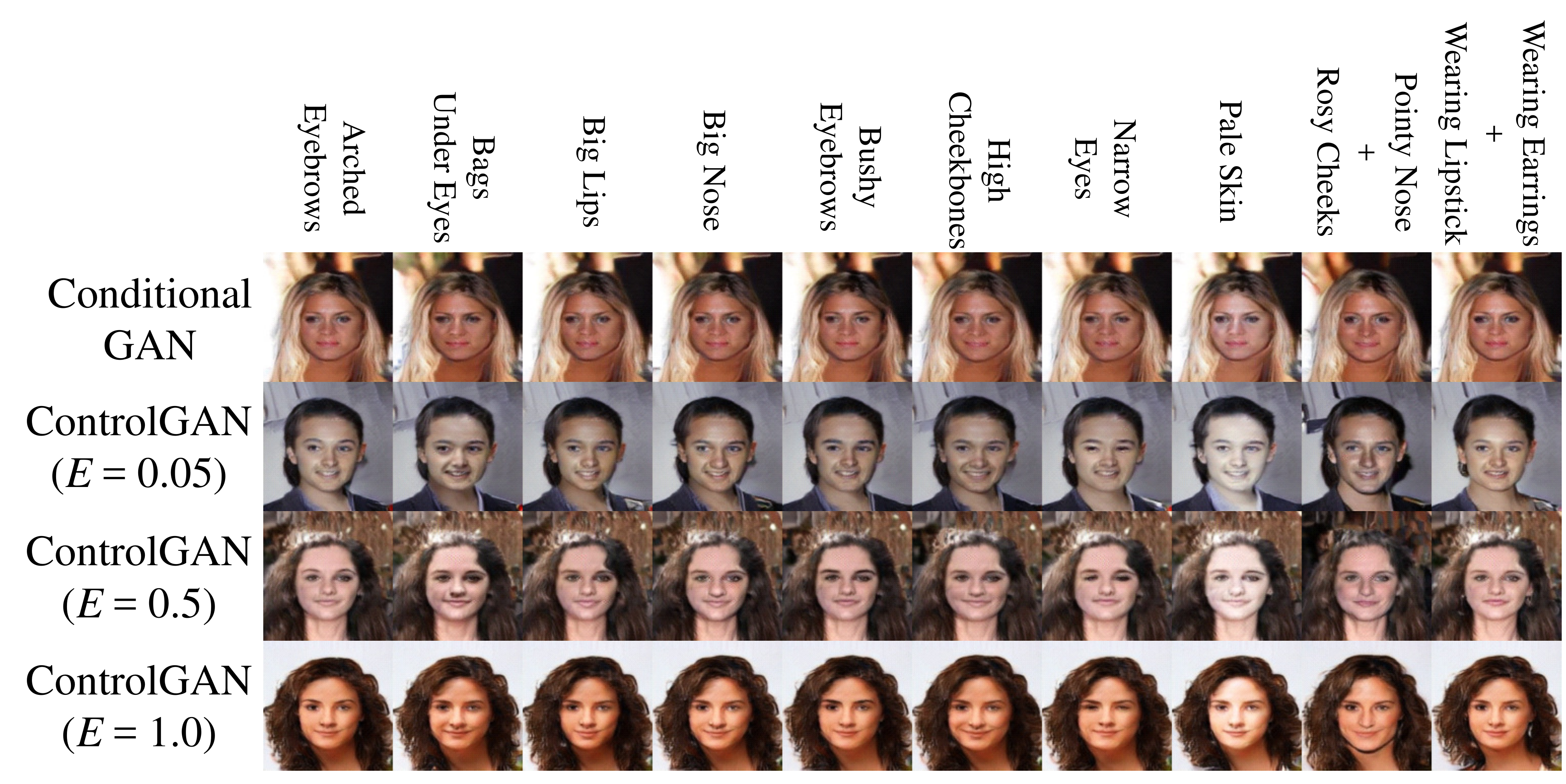}
    \caption{\textbf{Comparison between conditional GAN and ControlGAN with $E=0.05$, $0.5$ and $1.0$.} Images in each row are generated with a same input noise $z$. Each column denotes the corresponding labels. Two labels are used together in the last two column. All images are the size of $128 \times 128$.}
    \label{fig:fig_comp_add}
\end{sidewaysfigure}

\begin{sidewaysfigure}
    \centering
    \includegraphics[width=0.95\textwidth]{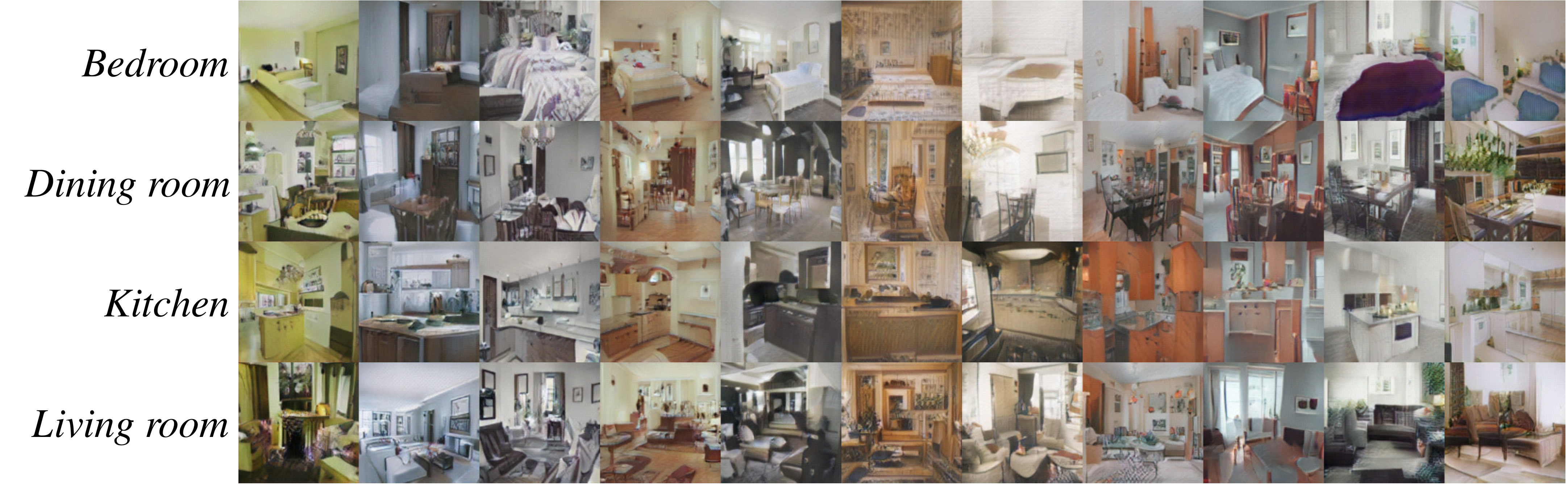}
    \caption{\textbf{$128 \times 128$ size of conditional room images generated by ControlGAN.} $E$ is set to 0.05. Note that the model was trained for one epoch.}
    \label{fig:fig_lsun_add}
\end{sidewaysfigure}

\begin{sidewaysfigure}
    \centering
    \includegraphics[width=0.95\textwidth]{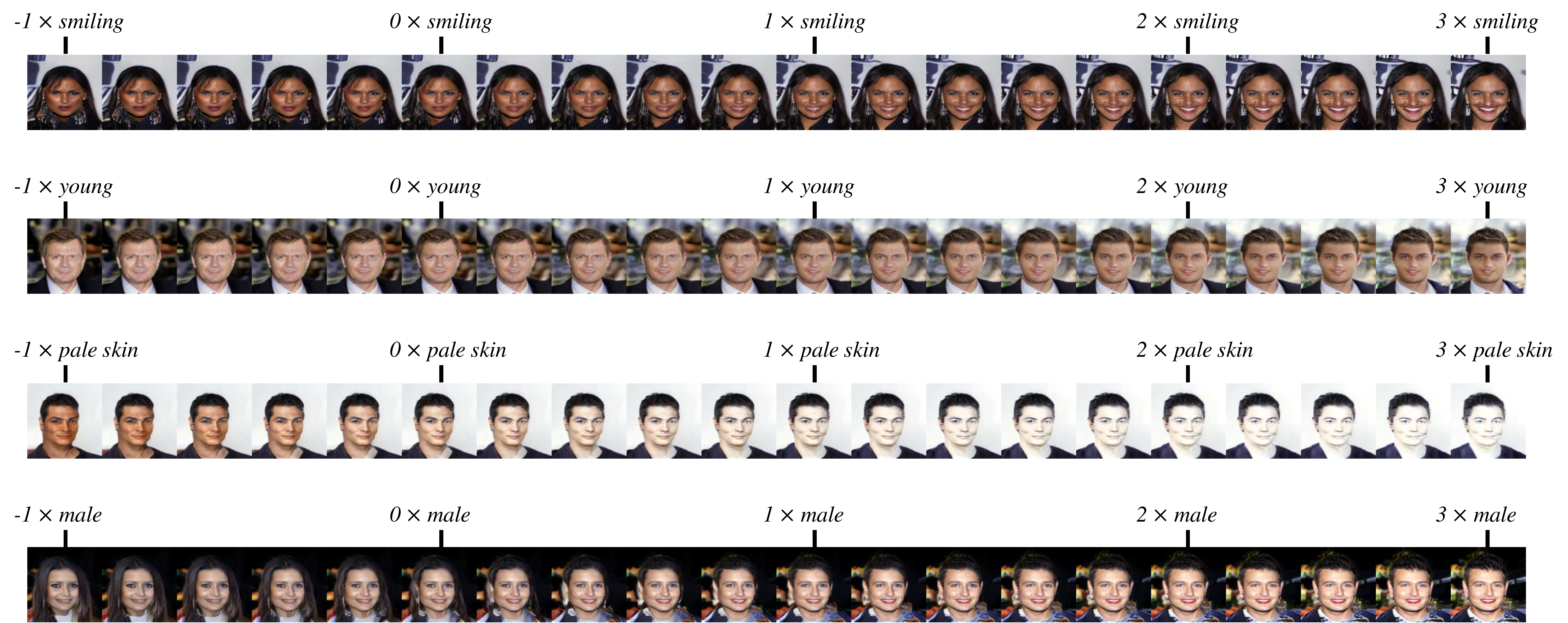}
    \caption{\textbf{Interpolation of the 'Smiling', 'Young', 'Pale Skin' and 'Male' labels.} Note that the model was trained only with $0 \times label$ and $1 \times label$. The values of $[-1.0, 0.0)$, $(0.0, 1.0)$ and $(1.0, 3.0]$ in this figure had never been trained in the training process. $E$ is set to 0.05.}
    \label{fig:fig_inter_add1}
\end{sidewaysfigure}
    
\begin{sidewaysfigure}
    \centering
    \includegraphics[width=0.95\textwidth]{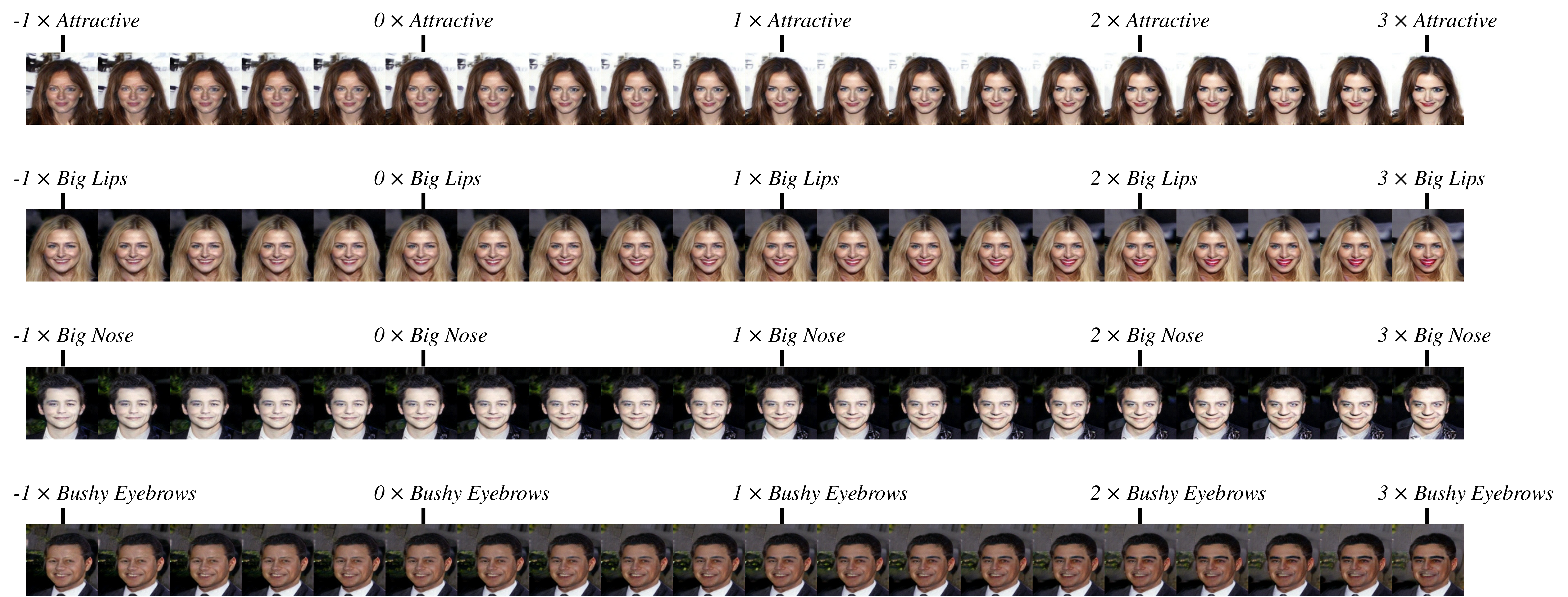}
    \caption{\textbf{Interpolation of the 'Attractive', 'Big Lips', 'Big Nose' and 'Bushy Eyebrows' labels.} Note that the model was trained only with $0 \times label$ and $1 \times label$. The values of $[-1.0, 0.0)$, $(0.0, 1.0)$ and $(1.0, 3.0]$ in this figure had never been trained in the training process. $E$ is set to 0.05.}
    \label{fig:fig_inter_add2}
\end{sidewaysfigure}

\begin{sidewaysfigure}
    \centering
    \includegraphics[width=0.95\textwidth]{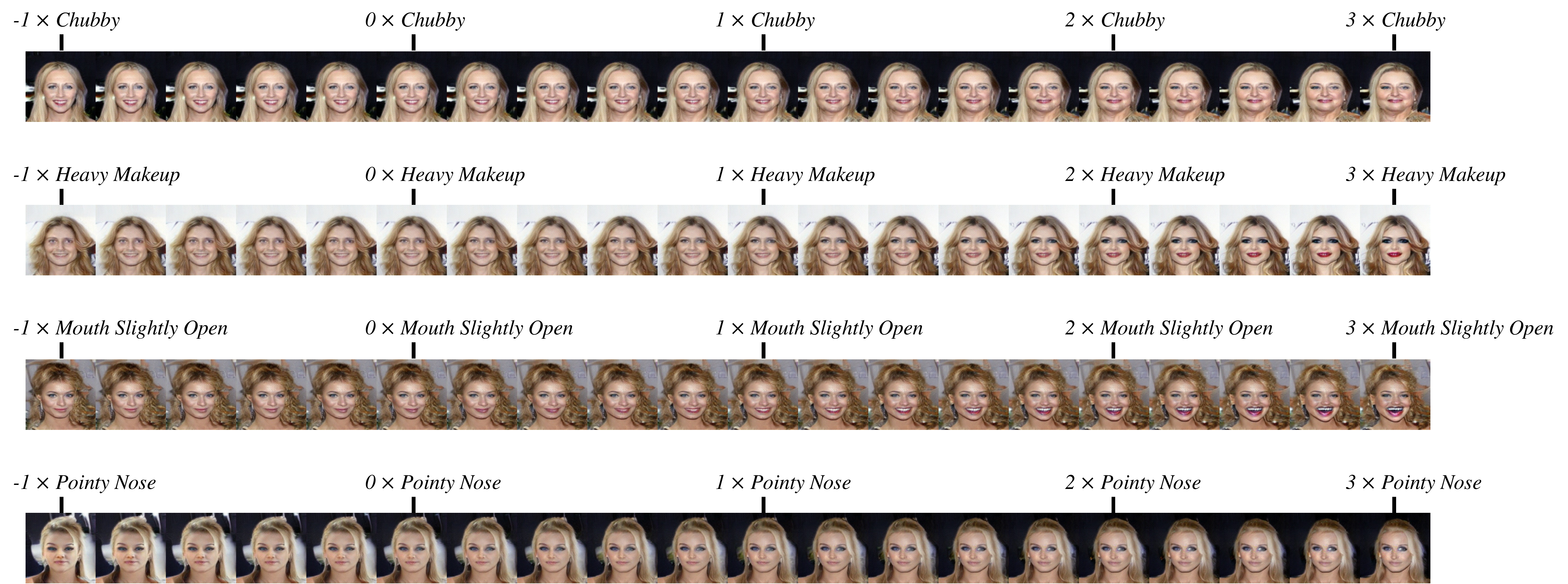}
    \caption{\textbf{Interpolation of the 'Chubby', 'Heavy Makeup', 'Mouth Slightly Open' and 'Pointy Nose' labels.} Note that the model was trained only with $0 \times label$ and $1 \times label$. The values of $[-1.0, 0.0)$, $(0.0, 1.0)$ and $(1.0, 3.0]$ in this figure had never been trained in the training process. $E$ is set to 0.05.}
    \label{fig:fig_inter_add3}
\end{sidewaysfigure}
\end{document}